\documentclass{article}

\usepackage[final]{neurips_2019}

\usepackage[utf8]{inputenc} 
\usepackage[T1]{fontenc}    
\usepackage{hyperref}       
\usepackage{url}            
\usepackage{booktabs}       
\usepackage{amsfonts}       
\usepackage{nicefrac}       
\usepackage{microtype}      
\usepackage{subfigure}
\usepackage{color}

\usepackage[numbers]{natbib} 

\usepackage{mathtools}
\mathtoolsset{showonlyrefs=true}

\usepackage{xargs}    
\usepackage[colorinlistoftodos,prependcaption,textsize=tiny]{todonotes}
\newcommandx{\unsure}[2][1=]{\todo[linecolor=red,backgroundcolor=red!25,bordercolor=red,#1]{#2}}
\newcommandx{\change}[2][1=]{\todo[linecolor=blue,backgroundcolor=blue!25,bordercolor=blue,#1]{#2}}
\newcommandx{\info}[2][1=]{\todo[linecolor=OliveGreen,backgroundcolor=OliveGreen!25,bordercolor=OliveGreen,#1]{#2}}
\newcommandx{\improvement}[2][1=]{\todo[linecolor=Plum,backgroundcolor=Plum!25,bordercolor=Plum,#1]{#2}}

\title{Convergent Policy Optimization for Safe Reinforcement Learning}

\author{
  Ming Yu \thanks{The University of Chicago Booth School of Business, Chicago, IL. Email: \texttt{ming93@uchicago.edu}. }\\
  \And
  Zhuoran Yang \thanks{Department of Operations Research and Financial Engineering, Princeton University, Princeton, NJ.}\\
  \And
  Mladen Kolar \thanks{The University of Chicago Booth School of Business, Chicago, IL.}\\
  \And
  Zhaoran Wang \thanks{Department of Industrial Engineering and Management Sciences, Northwestern University, Evanston, IL.} 
  }

\usepackage{mystyle}
\graphicspath{{figs/}}

\begin{document}

\maketitle

\begin{abstract}%
  We study the safe reinforcement learning problem with nonlinear
  function approximation, where policy optimization is formulated as a
  constrained optimization problem with both the objective and the
  constraint being nonconvex functions. For such a problem, we
  construct a sequence of surrogate convex constrained optimization
  problems by replacing the nonconvex functions locally with convex
  quadratic functions obtained from policy gradient estimators.  We
  prove that the solutions to these surrogate problems converge to a
  stationary point of the original nonconvex problem. Furthermore, to
  extend our theoretical results, we apply our algorithm to examples
  of optimal control and multi-agent reinforcement learning with
  safety constraints.
\end{abstract}


\section{Introduction}
\label{sec:introduction}

Reinforcement learning \citep{sutton2018reinforcement} has achieved
tremendous success in video games \citep{mnih2015human,
  peng2017multiagent, sun2018tstarbots, lee2018modular,xu2018macro}
and board games, such as chess and Go \citep{silver2016mastering,
  silver2017mastering, silver2018general}, in part due to powerful
simulators \citep{bellemare2013arcade, vinyals2017starcraft}.  In
contrast, due to physical limitations, real-world applications of
reinforcement learning methods often need to take into consideration
the safety of the agent \citep{amodei2016concrete,
  garcia2015comprehensive}.  For instance, in expensive robotic and
autonomous driving platforms, it is pivotal to avoid damages and
collisions \citep{fisac2018general, berkenkamp2017safe}.
In medical applications, we need to consider the switching cost \cite{bai2019provably}.
 
A popular model of safe reinforcement learning is the constrained
Markov decision process (CMDP), which generalizes the Markov decision
process by allowing for inclusion of constraints that model the
concept of safety \citep{altman1999constrained}.  In a CMDP, the cost is
associated with each state and action experienced by the agent, and
safety is ensured only if the expected cumulative cost is below a
certain threshold.  Intuitively, if the agent takes an unsafe action
at some state, it will receive a huge cost that punishes risky
attempts.  Moreover, by considering the cumulative cost, the notion of
safety is defined for the whole trajectory enabling us to examine the
long-term safety of the agent, instead of focusing on individual
state-action pairs. For a CMDP, the goal is to take sequential
decisions to achieve the expected cumulative reward under the safety
constraint.

Solving a CMDP can be written as a linear program
\cite{altman1999constrained}, with the number of variables being the
same as the size of the state and action spaces. Therefore, such an
approach is only feasible for the tabular setting, where we can
enumerate all the state-action pairs. For large-scale reinforcement
learning problems, where function approximation is applied, both the
objective and constraint of the CMDP are nonconvex functions of the
policy parameter.  One common method for solving CMDP is to formulate
an unconstrained saddle-point optimization problem via Lagrangian
multipliers and solve it using policy optimization algorithms
\cite{chow2017risk, tessler2018reward}. Such an approach suffers the
following two drawbacks:

First, for each fixed Lagrangian multiplier, the inner minimization
problem itself can be viewed as solving a new reinforcement learning
problem. From the computational point of view, solving the
saddle-point optimization problem requires solving a sequence of MDPs
with different reward functions. For a large scale problem, even
solving a single MDP requires huge computational resources, making
such an approach computationally infeasible.

Second, from a theoretical perspective, the performance of the
saddle-point approach hinges on solving the inner problem optimally.
Existing theory only provides convergence to a stationary point where
the gradient with respect to the policy parameter is zero
\citep{grondman2012survey, li2017deep}.  Moreover, the objective, as a
bivariate function of the Lagrangian multiplier and the policy
parameter, is not convex-concave and, therefore, first-order iterative
algorithms can be unstable \citep{goodfellow2014generative}.

In contrast, we tackle the nonconvex constrained optimization problem
of the CMDP directly. We propose a novel policy optimization
algorithm, inspired by \cite{liu2018stochastic}. Specifically, in each
iteration, we replace both the objective and constraint by quadratic
surrogate functions and update the policy parameter by solving the new
constrained optimization problem.  The two surrogate functions can be
viewed as first-order Taylor-expansions of the expected reward and
cost functions where the gradients are estimated using policy gradient
methods \citep{sutton2000policy}.  Additionally, they can be viewed as
convex relaxations of the original nonconvex reward and cost
functions. In \S\ref{sec:theoretical} we show that, as the algorithm
proceeds, we obtain a sequence of convex relaxations that gradually
converge to a smooth function. More importantly, the sequence of
policy parameters converges almost surely to a stationary point of the
nonconvex constrained optimization problem.

\paragraph{Related work.} 
Our work is pertinent to the line of research on CMDP
\citep{altman1999constrained}. For CMDPs with large state and action
spaces, \cite{chow2018lyapunov} proposed an iterative algorithm based
on a novel construction of Lyapunov functions. However, their theory
only holds for the tabular setting.  Using Lagrangian multipliers,
\citep{prashanth2016variance,chow2017risk,achiam2017constrained,
  tessler2018reward} proposed policy gradient
\citep{sutton2000policy}, actor-critic \citep{konda2000actor}, or
trust region policy optimization \citep{schulman2015trust} methods for
CMDP or constrained risk-sensitive reinforcement learning
\citep{garcia2015comprehensive}.  These algorithms either do not have
convergence guarantees or are shown to converge to saddle-points of
the Lagrangian using two-time-scale stochastic approximations
\citep{borkar1997stochastic}. However, due to the projection on the
Lagrangian multiplier, the saddle-point achieved by these approaches
might not be the stationary point of the original CMDP problem. In
addition, \cite{wen2018constrained} proposed a cross-entropy-based
stochastic optimization algorithm, and proved the asymptotic behavior
using ordinary differential equations.  In contrast, our algorithm and
the theoretical analysis focus on the discrete time CMDP.  Outside of
the CMDP setting, \citep{huang2018learning, lacotte2018risk} studied
safe reinforcement learning with demonstration data,
\cite{turchetta2016safe} studied the safe exploration problem with
different safety constraints, and \cite{ammar2015safe} studied
multi-task safe reinforcement learning.

\paragraph{Our contribution.} 
Our contribution is three-fold.  First, for the CMDP policy
optimization problem where both the objective and constraint function
are nonconvex, we propose to optimize a sequence of convex relaxation
problems using convex quadratic functions. Solving these surrogate
problems yields a sequence of policy parameters that converge almost
surely to a stationary point of the original policy optimization
problem. Second, to reduce the variance in the gradient estimator that
is used to construct the surrogate functions, we propose an online
actor-critic algorithm.  Finally, as concrete applications, our
algorithms are also applied to optimal control (in
\S\ref{sec:application_LQR}) and parallel and multi-agent
reinforcement learning problems with safety constraints (in
supplementary material).



\vspace{-1mm}
\section{Background}
\label{sec:background}

A Markov decision process is denoted by
$(\cS, \cA, P, \gamma, r, \mu)$, where $\cS$ is the state space, $\cA$
is the action space, $P$ is the transition probability distribution,
$\gamma \in (0,1)$ is the discount factor,
$r\colon \cS \times \cA \rightarrow \RR$ is the reward function, and
$\mu \in \cP(\cS)$ is the distribution of the initial state
$s_0\in \cS$, where we denote $\cP(\cX)$ as the set of probability
distributions over $\cX$ for any $\cX$.  A policy is a mapping
$\pi: \cS \rightarrow \cP(\cA)$ that specifies the action that an agent
will take when it is at state $s$.

\vspace{-1mm}
\paragraph{Policy gradient method.}

Let $\{ \pi_{\theta} \colon \cS \rightarrow \cP(\cA) \}$ be a
parametrized policy class, where $\theta \in \Theta$ is the parameter
defined on a compact set $\Theta$. This parameterization transfers the
original infinite dimensional policy class to a finite dimensional
vector space and enables gradient based methods to be used to maximize
\eqref{eq:expected_reward}.  For example, the most popular Gaussian
policy can be written as
$\pi(\cdot | s, \theta) = \cN \big(\mu(s, \theta), \sigma(s, \theta)
\big)$, where the state dependent mean $\mu(s, \theta)$ and standard
deviation $\sigma(s, \theta)$ can be further parameterized as
$\mu(s, \theta) = \theta_{\mu}^\top \cdot x(s)$ and
$\sigma(s, \theta) = \exp\big(\theta_{\sigma}^\top \cdot x(s)\big)$
with $x(s)$ being a state feature vector.  The goal of an agent is to
maximize the expected cumulative reward
\begin{equation}
  \label{eq:expected_reward}
  R(\theta) = \EE_\pi
  \biggl [ \sum_{t\geq 0} \gamma ^t \cdot r(s_t, a_t) \biggr ],  
\end{equation}
where $s_0 \sim \mu$, and for all $t\geq 0$, we have
$s_{t+1} \sim P(\cdot \given s_t, a_t)$ and
$a_t \sim \pi(\cdot \given s_t)$.
Given a policy $\pi(\theta)$, we define the state- and action-value functions of $\pi_{\theta}$, respectively, as 
\#
\label{eq:value_funcs}
V^{\theta} (s) = \EE_{\pi_{\theta} } \biggl [ \sum_{t\geq 0} \gamma ^t r(s_t, a_t) \bigggiven s_0 = s \biggr ],  \
Q^{\theta} (s,a) = \EE_{\pi_{\theta} } \biggl [ \sum_{t\geq 0} \gamma ^t r(s_t, a_t) \bigggiven s_0 = s, a_0 = a\biggr ].
\#
The policy gradient method updates the parameter $\theta$ through gradient ascent
\#
\theta_{k+1} = \theta_{k} + \eta \cdot \hat{\nabla}_{\theta} R(\theta_k),
\#
where $\hat{\nabla}_{\theta} R(\theta_k)$ is a stochastic estimate of
the gradient $\nabla _{\theta} R(\theta_k)$ at $k$-th iteration.  
Policy gradient method, as well as its variants (e.g. policy gradient with baseline \cite{sutton2018reinforcement}, neural policy gradient \cite{wang2019neural, liu2019neural, cai2019neural}) is widely used in reinforcement learning.
The gradient
$\nabla _{\theta} R(\theta)$ can be 
estimated according to the policy gradient
theorem \citep{sutton2000policy},
\#
\label{eq:pg_theorem}
\nabla_{\theta} R(\theta) = \EE \Big[ \nabla_{\theta} \log \pi_{\theta}(s,a) \cdot Q^{\theta}(s,a) \Big].
\#

\vspace{-1mm}
\paragraph{Actor-critic method.}

To further reduce the variance of the policy gradient method, we could
estimate both the policy parameter and value function
simultaneously. This kind of method is called actor-critic algorithm
\cite{konda2000actor}, which is widely used in reinforcement
learning. Specifically, in the value function evaluation
(\emph{critic}) step we estimate the action-value function
$Q^{\theta}(s,a)$ using, for example, the temporal difference method
TD(0) \cite{dann2014policy}. The policy parameter update
(\emph{actor}) step is implemented as before by the Monte-Carlo method
according to the policy gradient theorem \eqref{eq:pg_theorem} with
the action-value $Q^{\theta}(s,a)$ replaced by the estimated value in
the policy evaluation step.

\vspace{-1mm}
\paragraph{Constrained MDP.}

In this work, we consider an MDP problem with an additional constraint
on the model parameter $\theta$. Specifically, when taking action at
some state we incur some cost value. The constraint is such that the
expected cumulative cost cannot exceed some pre-defined constant.  A
constrained Markov decision process (CMDP) is denoted by
$(\cS, \cA, P, \gamma, r, d, \mu)$, where
$d\colon \cS \times \cA \rightarrow \RR$ is the cost function and the
other parameters are as before.  The goal of an the agent in CMDP is
to solve the following constrained problem
\begin{equation}
\begin{aligned}
\label{eq:constrained}
&\mathop{\textrm{minimize}}_{\theta \in \Theta} 
~~J(\theta) =    \EE_{\pi_{\theta} } \biggl [ - \sum_{t\geq 0} \gamma ^t \cdot r(s_t, a_t) \biggr ] , \\
&\text{subject to}
~~D(\theta) =   \EE_{\pi_{\theta} } \biggl [  \sum_{t\geq 0} \gamma ^t \cdot  d(s_t, a_t) \biggr ]  \leq  D_0,
\end{aligned}
\end{equation}
where $D_0 $ is a fixed constant. We consider only one constraint
$D(\theta) \leq D_0$, noting that it is straightforward to generalize
to multiple constraints. Throughout this paper, we assume that both
the reward and cost value functions are bounded:
$\big|r(s_t, a_t)\big| \leq r_{\max}$ and
$\big|d(s_t, a_t)\big| \leq d_{\max}$. Also, the parameter space
$\Theta$ is assumed to be compact.


\vspace{-1mm}
\section{Algorithm}
\label{sec:algorithm}

In this section, we develop an algorithm to solve the optimization
problem \eqref{eq:constrained}.  Note that both the objective function
and the constraint in \eqref{eq:constrained} are nonconvex and involve
expectation without closed-form expression.  As a constrained problem,
a straightforward approach to solve \eqref{eq:constrained} is to
define the following Lagrangian function
\#
L(\theta, \lambda) = J(\theta) + \lambda \cdot \big[ D(\theta) - D_0 \big],
\#
and solve the dual problem
\#
\inf_{\lambda \geq 0}  \sup_{\theta} L(\theta, \lambda).
\#
However, this problem is a nonconvex minimax problem and, therefore,
is hard to solve and establish theoretical guarantees for solutions
\cite{adolphs2018non}.  Another approach to solve
\eqref{eq:constrained} is to replace $J(\theta)$ and $D(\theta)$ by
surrogate functions with nice properties.  For example, one can
iteratively construct local quadratic approximations that are strongly
convex \cite{scutari2013decomposition}, or are an upper bound for the
original function \cite{sun2016majorization}.  However, an immediate
problem of this naive approach is that, even if the original problem
\eqref{eq:constrained} is feasible, the convex relaxation problem need
not be.  Also, these methods only deal with deterministic and/or
convex constraints.

In this work, we propose an iterative algorithm that approximately
solves \eqref{eq:constrained} by constructing a sequence of convex
relaxations, inspired by \cite{liu2018stochastic}.  Our method is able
to handle the possible infeasible situation due to the convex
relaxation as mentioned above, and handle stochastic and nonconvex
constraint.
Since we do not have access to $J(\theta)$ or $D(\theta)$, we first
define the sample negative cumulative reward and cost functions as
\#
J^*(\theta) = - \sum_{t\geq 0} \gamma ^t \cdot  r(s_t, a_t) 
\qquad \text{and}\qquad
D^*(\theta) = \sum_{t\geq 0} \gamma ^t \cdot  d(s_t, a_t).
\#
Given $\theta$, $J^*(\theta)$ and $D^*(\theta)$ are the sample
negative cumulative reward and cost value of a realization (i.e., a
trajectory) following policy $\pi_{\theta}$.  Note that both
$J^*(\theta)$ and $D^*(\theta)$ are stochastic due to the randomness
in the policy, state transition distribution, etc.  With some abuse of
notation, we use $J^*(\theta)$ and $D^*(\theta)$ to
denote both a function of $\theta$ and a value obtained by the
realization of a trajectory.  Clearly we have
$J(\theta) = \EE \big[ J^*(\theta) \big]$ and
$D(\theta) = \EE \big[ D^*(\theta) \big]$.

We start from some (possibly infeasible) $\theta_0$.  Let $\theta_k$
denote the estimate of the policy parameter in the $k$-th iteration.
As mentioned above, we do not have access to the expected cumulative
reward $J(\theta)$. Instead we sample a trajectory following the
current policy $\pi_{\theta_k}$ and obtain a realization of the
negative cumulative reward value and the gradient of it as
$J^*(\theta_k)$ and $\nabla_{\theta} J^*(\theta_k)$, respectively. The
cumulative reward value is obtained by Monte-Carlo estimation, and the
gradient is also obtained by Monte-Carlo estimation according to the
policy gradient theorem in \eqref{eq:pg_theorem}.  We provide more
details on the realization step later in this section.  Similarly, we
use the same procedure for the cost function and obtain realizations
$D^*(\theta_k)$ and $\nabla_{\theta} D^*(\theta_k)$.

We approximate $J(\theta)$ and $D(\theta)$ at $\theta_k$ by the quadratic surrogate functions
\#
\label{eq:quadratic_appriximation_J}
\tilde J(\theta, \theta_k, \tau ) & = J^*(\theta_k) + \la \nabla_{\theta} J^*(\theta_k), \theta - \theta_k \ra + \tau \| \theta - \theta_k \|_2^2, \\
\label{eq:quadratic_appriximation_D}
\tilde D(\theta, \theta_k, \tau ) &= D^*(\theta_k)  +  \la \nabla _{\theta} D^*(\theta_k), \theta - \theta_k \ra + \tau  \| \theta - \theta_k \|_2^2,
\# 
where $\tau > 0$ is any fixed constant.
In each iteration, we solve the optimization problem 
\begin{align}
\label{eq:QCQP}
\overline \theta_k =  
  \argmin _{\theta }  \overline J^{(k)} (\theta)   \qquad 
  \text{subject to} \qquad  \overline  D^{(k)} (\theta) \leq D_0,
\end{align}
where we define  
\#
\label{eq:J_bar}
\overline J^{(k)} (\theta) = (1 - \rho_k) \cdot \overline J^{(k-1)} (\theta) + \rho_k \cdot \tilde J(\theta, \theta_k, \tau), \\
\label{eq:D_bar}
\overline D^{(k)} (\theta) = (1 - \rho_k) \cdot \overline D^{(k-1)} (\theta) + \rho_k \cdot \tilde D(\theta, \theta_k, \tau),
\# 
with the initial value
$\overline J^{(0)}(\theta) = \overline D^{(0)}(\theta) = 0$. Here
$\rho_k$ is the weight parameter to be specified later.  According to
the definition \eqref{eq:quadratic_appriximation_J} and
\eqref{eq:quadratic_appriximation_D}, problem \eqref{eq:QCQP} is a
convex quadratically constrained quadratic program (QCQP). Therefore,
it can be efficiently solved by, for example, the interior
point method.  However, as mentioned before, even if the original
problem \eqref{eq:constrained} is feasible, the convex relaxation
problem \eqref{eq:QCQP} could be infeasible. In this case, we instead
solve the following feasibility problem
\begin{equation}
\begin{aligned}
\label{eq:QCQP_alternative}
\overline \theta_k = ~
 \argmin _{\theta, \alpha } ~~ \alpha \qquad
 \text{subject to} \qquad \overline  D^{(k)} (\theta) \leq D_0 + \alpha.
\end{aligned}
\end{equation}
In particular, we relax the infeasible constraint and find
$\overline \theta_k$ as the solution that gives the minimum
relaxation.  Due to the specific form in~\eqref{eq:quadratic_appriximation_D},
$\overline D^{(k)} (\theta)$ is decomposable into quadratic forms of each component of $\theta$,
with no terms involving $\theta_i \cdot \theta_j$. Therefore, the
solution to problem \eqref{eq:QCQP_alternative} can be written in a
closed form.  Given $\overline \theta_k$ from either \eqref{eq:QCQP}
or \eqref{eq:QCQP_alternative}, we update $\theta_k$ by
\#
\label{eq:theta_update}
\theta_{k+1} = ( 1- \eta_k) \cdot \theta_k + \eta_k \cdot \overline \theta_k, 
\# 
where $\eta_k$ is the learning rate to be specified later. 
Note that although we consider only one constraint in the algorithm,
both the algorithm and the theoretical result in Section
\ref{sec:theoretical} can be directly generalized to multiple
constraints setting.  The whole procedure is summarized in Algorithm
\ref{algo}.

\begin{algorithm}[tb]
   \caption{Successive convex relaxation algorithm for constrained MDP}
   \label{algo}
\begin{algorithmic}[1]
   \STATE {\bfseries Input:} Initial value $\theta_0$, $\tau$, $\{ \rho_k \}, \{ \eta_k \}$.
   \FOR{$k=1, 2, 3, \dots$}
   \STATE Obtain a sample $J^*(\theta_k)$ and $D^*(\theta_k)$ by Monte-Carlo sampling.
   \STATE Obtain a sample $\nabla _{\theta} J^*(\theta_k)$ and $\nabla _{\theta} D^*(\theta_k)$ by policy gradient theorem.
       \IF {problem \eqref{eq:QCQP} is feasible}
        \STATE Obtain $\overline \theta_k$ by solving \eqref{eq:QCQP}.
    \ELSE
        \STATE Obtain $\overline \theta_k$ by solving \eqref{eq:QCQP_alternative}.
    \ENDIF
   \STATE Update $\theta_{k+1}$ by \eqref{eq:theta_update}.
   \ENDFOR
\end{algorithmic}
\end{algorithm}

\vspace{-1.5mm}
\paragraph{Obtaining realizations $J^*(\theta_k)$ and $\nabla_{\theta} J^*(\theta_k)$.}
We detail how to obtain realizations $J^*(\theta_k)$ and
$\nabla_{\theta} J^*(\theta_k)$ corresponding to the lines 3 and 4 in
Algorithm \ref{algo}. The realizations of $D^*(\theta_k)$ and
$\nabla_{\theta} D^*(\theta_k)$ can be obtained similarly.

First, we discuss finite horizon setting, where we can sample the full
trajectory according to the policy $\pi_{\theta}$. In particular, for
any $\theta_k$, we use the policy $\pi_{\theta_k}$ to sample a
trajectory and obtain $J^*(\theta_k)$ by Monte-Carlo method.  The
gradient $\nabla _{\theta} J(\theta)$ can be estimated by the policy
gradient theorem \cite{sutton2000policy},
\#
\label{eq:pg_theorem_algo}
\nabla_{\theta} J(\theta) = - \EE_{\pi_{\theta}} \Big[ \nabla_{\theta} \log \pi_{\theta}(s,a) \cdot Q^{\theta}(s,a) \Big].
\#
Again we can sample a trajectory and obtain the policy gradient
realization $\nabla _{\theta} J^*(\theta_k)$ by Monte-Carlo method.

In infinite horizon setting, we cannot sample the infinite length
trajectory. In this case, we utilize the truncation method introduced
in \cite{rhee2015unbiased}, which truncates the trajectory at some
stage $T$ and scales the undiscounted cumulative reward to obtain an
unbiased estimation.  Intuitively, if the discount factor $\gamma$ is
close to $0$, then the future reward would be discounted heavily and,
therefore, we can obtain an accurate estimate with a relatively small
number of stages. On the other hand, if $\gamma$ is close to $1$, then
the future reward is more important compared to the small $\gamma$
case and we have to sample a long trajectory. Taking this intuition
into consideration, we define $T$ to be a geometric random variable
with parameter $1 - \gamma$: $\Pr(T = t) = (1-\gamma)
\gamma^{t}$. Then, we simulate the trajectory until stage $T$ and use
the estimator
$J_{\text{truncate}}(\theta) = - (1-\gamma) \cdot \sum_{t = 0}^T r(s_t, a_t) $,
which is an unbiased estimator of the expected
negative cumulative reward $J(\theta)$, as proved in proposition 5 in \cite{paternain2018stochastic}. 
We can apply the same truncation procedure to estimate the policy
gradient $\nabla_{\theta} J(\theta)$.

\vspace{-1.5mm}
\paragraph{Variance reduction.} 

Using the naive sampling method described above, we may suffer from
high variance problem.  To reduce the variance, we can modify the
above procedure in the following ways.  First, instead of sampling
only one trajectory in each iteration, a more practical and stable way
is to sample several trajectories and take average to obtain the
realizations. As another approach, we can subtract a baseline
function from the action-value function $Q^{\theta}(s,a)$ in the
policy gradient estimation step \eqref{eq:pg_theorem_algo} to reduce
the variance without changing the expectation. A popular choice of
the baseline function is the state-value function $V^{\theta} (s)$ as
defined in \eqref{eq:value_funcs}. In this way, we can replace
$Q^{\theta}(s,a)$ in \eqref{eq:pg_theorem_algo} by the advantage
function $A^\theta(s, a)$ defined as
\begin{equation*}
  A^\theta(s, a) =
Q^{\theta}(s,a) - V^{\theta} (s).
\end{equation*}
This modification corresponds to the standard REINFORCE with Baseline
algorithm \cite{sutton2018reinforcement} and can significantly reduce
the variance of policy gradient.

\vspace{-1.5mm}
\paragraph{Actor-critic method.}
Finally, we can use an actor-critic update to improve the
performance further.  In this case, since we need unbiased estimators for both the gradient
and the reward value in \eqref{eq:quadratic_appriximation_J} and
\eqref{eq:quadratic_appriximation_D} in online fashion, we modify our original problem
\eqref{eq:constrained} to average reward setting as
\begin{equation}
\begin{aligned}
\label{eq:constrained_average}
&\mathop{\textrm{minimize}}_{\theta \in \Theta} 
~~J(\theta) =  \lim_{T \to \infty}  \EE_{\pi_{\theta} } \biggl [ - \frac{1}{T} \sum_{t = 0}^T r(s_t, a_t) \biggr ],  \\
&\text{subject to}
~~D(\theta) =  \lim_{T \to \infty} \EE_{\pi_{\theta} } \biggl [ \frac{1}{T} \sum_{t = 0}^T d(s_t, a_t) \biggr ]  \leq  D_0.
\end{aligned}
\end{equation}
Let $V^J_\theta(s)$ and $V^D_\theta(s)$ denote the value and cost
functions corresponding to \eqref{eq:value_funcs}.  We use possibly
nonlinear approximation with parameter $w$ for the value function:
$V^J_w (s)$ and $v$ for the cost function: $V^D_v (s)$.  In the critic
step, we update $w$ and $v$ by TD(0) with step size $\beta_w$ and
$\beta_v$; in the actor step, we solve our proposed convex relaxation
problem to update $\theta$.  The actor-critic procedure is summarized
in Algorithm \ref{algo_ac}.  Here $J$ and $D$ are estimators of
$J(\theta_k)$ and $D(\theta_k)$. Both of $J$ and $D$, and the TD error
$\delta^J$, $\delta^D$ can be initialized as 0.

The usage of the actor-critic method helps reduce variance by using a
value function instead of Monte-Carlo sampling. Specifically, in
Algorithm \ref{algo} we need to obtain a sample trajectory and
calculate $J^*(\theta)$ and $\nabla_\theta J^*(\theta)$ by Monte-Carlo
sampling. This step has a high variance since we need to sample a
potentially long trajectory and sum up a lot of random rewards.  In
contrast, in Algorithm \ref{algo_ac}, this step is replaced by a value
function $V^J_w(s)$, which reduces the variance.

\begin{algorithm}[tb]
   \caption{Actor-Critic update for constrained MDP}
   \label{algo_ac}
\begin{algorithmic}[1]
   \FOR{$k=1, 2, 3, \dots$ }
   \STATE Take action $a$, observe reward $r$, cost $d$, and new state $s'$.
   \STATE {\bf Critic step:}
   \STATE ~~~ $w \leftarrow w + \beta_w \cdot \delta^J \nabla_w V^J_w (s), ~~ J \leftarrow J + \beta_w \cdot\big( r - J \big)$.
   \STATE ~~~ $v \leftarrow v + \beta_v \cdot \delta^D \nabla_v V^J_v (s), ~~ D \leftarrow D + \beta_v \cdot\big( d - D \big)$.
   \STATE {\bf Calculate TD error:}
   \STATE ~~~ $\delta^J = r - J + V^J_w(s') - V^J_w(s)$.
   \STATE ~~~ $\delta^D = d - D + V^D_v(s') - V^D_v(s)$.
   \STATE {\bf Actor step:}
   \STATE ~~~ Solve $\overline \theta_k$ by \eqref{eq:QCQP} or \eqref{eq:QCQP_alternative} with \\
   \qquad $J^*(\theta_k)$, $\nabla _{\theta} J^*(\theta_k)$ in \eqref{eq:quadratic_appriximation_J} replaced by $J$ and $\delta^J \cdot \nabla_{\theta} \log \pi_{\theta}(s,a) $; \\
   \qquad $D^*(\theta_k)$, $\nabla _{\theta} D^*(\theta_k)$ in \eqref{eq:quadratic_appriximation_D} replaced by $D$ and $\delta^D \cdot \nabla_{\theta} \log \pi_{\theta}(s,a) $.
   \STATE $s \leftarrow s'$.
   \ENDFOR
\end{algorithmic}
\end{algorithm}



\section{Theoretical Result}
\label{sec:theoretical}

In this section, we show almost sure convergence of the iterates
obtained by our algorithm to a stationary point.  We start by stating
some mild assumptions on the original problem \eqref{eq:constrained}
and the choice of some parameters in Algorithm \ref{algo}.

\begin{assumption}
\label{assumption:step_size}
The choice of $\{ \eta_k \}$ and $\{ \rho_k \}$ satisfy 
$\lim_{k \to \infty} \sum_k \eta_k = \infty$, $\lim_{k \to \infty} \sum_k \rho_k = \infty$
and
$\lim_{k \to \infty} \sum_k \eta_k^2 + \rho_k^2 < \infty$.
Furthermore, we have $\lim_{k \to \infty} \eta_k / \rho_k  = 0$ and $\eta_k$ is decreasing.
\end{assumption}

\begin{assumption}
\label{assumption:bounded}
For any realization, $J^*(\theta)$ and $D^*(\theta)$ are continuously
differentiable as functions of $\theta$.  Moreover, $J^*(\theta)$,
$D^*(\theta)$, and their derivatives are uniformly Lipschitz
continuous.
\end{assumption}

Assumption \ref{assumption:step_size} allows us to specify the
learning rates.  A practical choice would be $\eta_k = k^{-c_1}$ and
$\rho_k = k^{-c_2}$ with $0.5 < c_2 < c_1 < 1$. This assumption is
standard for gradient-based algorithms.  Assumption
\ref{assumption:bounded} is also standard and is known to hold for a
number of models.  It ensures that the reward and cost functions are
sufficiently regular.  In fact, it can be relaxed such that each
realization is Lipschitz (not uniformly), and the event that we keep
generating realizations with monotonically increasing Lipschitz
constant is an event with probability 0.  See condition iv) in
\cite{yang2016parallel} and the discussion thereafter.  Also, see
\cite{pirotta2015policy} for sufficient conditions such that both the
expected cumulative reward function and the gradient of it are
Lipschitz.

The following Assumption \ref{assumption:feasible} is useful only when
we initialize with an infeasible point. We first state it here and we
will discuss this assumption after the statement of the main theorem.

\begin{assumption}
\label{assumption:feasible}
Suppose $(\theta_S, \alpha_S)$ is a stationary point of the optimization problem
\begin{equation}
\begin{aligned}
\label{eq:relaxation_infeasible}
 \mathop{\textrm{minimize}}_{\theta, \alpha} ~~ \alpha \qquad
 \text{subject to} \qquad D(\theta) \leq D_0 + \alpha.
\end{aligned}
\end{equation}
We have that $\theta_S$ is a feasible point of the original problem
\eqref{eq:constrained}, i.e. $D(\theta_S) \leq D_0$.
\end{assumption}

We are now ready to state the main theorem. 
\begin{theorem}
\label{thm:main}
Suppose the Assumptions \ref{assumption:step_size} and
\ref{assumption:bounded} are satisfied with small enough initial step size $\eta_0$.
Suppose also that, either $\theta_0$ is a feasible point, or
Assumption \ref{assumption:feasible} is satisfied.  If there is a
subsequence $\{ \theta_{k_j} \}$ of $\{ \theta_k \}$ that converges to
some $\tilde\theta$, then there exist uniformly continuous functions
$\hat J(\theta)$ and $\hat D(\theta)$ satisfying
\begin{equation*}
  \lim_{j \to
  \infty} \overline J^{(k_j)} (\theta) = \hat J(\theta) \qquad
\text{and}\qquad \lim_{j \to \infty} \overline D^{(k_j)} (\theta) =
\hat D(\theta).
\end{equation*}
Furthermore, suppose there exists $\theta$ such
that $\hat D(\theta) < D_0$ (i.e. the Slater's condition holds), then
$\tilde\theta$ is a stationary point of the original problem
\eqref{eq:constrained} almost surely.
\end{theorem}

The proof of Theorem \ref{thm:main} is provided in the supplementary
material.

Note that Assumption \ref{assumption:feasible} is not necessary if we
start from a feasible point, or we reach a feasible point in the
iterates, which could be viewed as an initializer.  Assumption
\ref{assumption:feasible} makes sure that the iterates in Algorithm
\ref{algo} keep making progress without getting stuck at any
infeasible stationary point. A similar condition is assumed in
\cite{liu2018stochastic} for an infeasible initializer.  If it turns
out that $\theta_0$ is infeasible and Assumption
\ref{assumption:feasible} is violated, then the convergent point may
be an infeasible stationary point of \eqref{eq:relaxation_infeasible}.
In practice, if we can find a feasible point of the original problem,
then we proceed with that point. Alternatively, we could generate
multiple initializers and obtain iterates for all of them. As long as
there is a feasible point in one of the iterates, we can view this
feasible point as the initializer and Theorem \ref{thm:main} follows
without Assumption \ref{assumption:feasible}.  In our later
experiments, for every single replicate, we could reach a feasible
point, and therefore Assumption \ref{assumption:feasible} is not
necessary.

Our algorithm does not guarantee safe exploration during the training phase. 
Ensuring safety during learning is a more challenging problem. 
Sometimes even finding a feasible point is not straightforward, 
otherwise Assumption \ref{assumption:feasible} is not necessary.

Our proposed algorithm is inspired by \cite{liu2018stochastic}.
Compared to \cite{liu2018stochastic} which deals with an optimization
problem, solving the safe reinforcement learning problem is more
challenging.  We need to verify that the Lipschitz condition is
satisfied, and also the policy gradient has to be estimated (instead
of directly evaluated as in a standard optimization problem).  The
usage of the Actor-Critic algorithm reduces the variance of the sampling,
which is unique to Reinforcement learning.


\section{Application to Constrained Linear-Quadratic Regulator}
\label{sec:application_LQR}

We apply our algorithm to the linear-quadratic regulator (LQR), which
is one of the most fundamental problems in control theory.  In the LQR
setting, the state dynamic equation is linear, the cost function is
quadratic, and the optimal control theory tells us that the optimal
control for LQR is a linear function of the state
\cite{evans2005introduction, anderson2007optimal}.  LQR can be viewed
as an MDP problem and it has attracted a lot of attention in the
reinforcement learning literature \cite{bradtke1993reinforcement,
  bradtke1994adaptive, dean2017sample, recht2018tour}.

We consider the infinite-horizon, discrete-time LQR problem.  Denote
$x_t$ as the state variable and $u_t$ as the control variable. The
state transition and the control sequence are given by
\begin{equation}
  \label{eq:LQR}
  \begin{aligned}
    x_{{t+1}}&=Ax_{t}+Bu_{t} + v_t,\\
    u_t &= - F x_t + w_t,    
  \end{aligned}
\end{equation}
where $v_t$ and $w_t$ represent possible Gaussian white noise, and the
initial state is given by $x_0$. The goal is to find the control
parameter matrix $F$ such that the expected total cost is minimized.
The usual cost function of LQR corresponds to the negative reward in
our setting and we impose an additional quadratic constraint on the
system. The overall optimization problem is given by
\begin{equation}
\begin{aligned}
\label{eq:LQR_constrained}
&\mathop{\textrm{minimize}}_{} 
~~J(F) =    \EE \biggl [ \sum_{t\geq 0} x_t^\top Q_1 x_t + u_t^\top R_1 u_t \biggr ], \\
&\text{subject to}
~~D(F) =   \EE \biggl [  \sum_{t\geq 0} x_t^\top Q_2 x_t + u_t^\top R_2 u_t \biggr ]  \leq  D_0,
\end{aligned}
\end{equation}
where $Q_1, Q_2, R_1,$ and $R_2$ are positive definite matrices. Note
that even thought the matrices are positive definite, both the
objective function $J$ and the constraint $D$ are nonconvex with
respect to the parameter $F$.  Furthermore, with the additional
constraint, the optimal control sequence may no longer be linear in
the state $x_t$. Nevertheless, in this work, we still consider linear
control given by \eqref{eq:LQR} and the goal is to find the best
linear control for this constrained LQR problem.  We assume that the
choice of $A, B$ are such that the optimal cost is finite.

\paragraph{Random initial state.}
We first consider the setting where the initial state $x_0 \sim \cD$
follows a random distribution $\cD$, while both the state transition
and the control sequence \eqref{eq:LQR} are deterministic (i.e.
$v_t = w_t = 0$). In this random initial state setting,
\cite{fazel2018global} showed that without the constraint, the policy
gradient method converges efficiently to the global optima in polynomial time.  In the constrained
case, we can explicitly write down the objective and constraint
function, since the only randomness comes from the initial state.
Therefore, we have the state dynamic $x_{t+1} = (A-BF)x_t$ and
the objective function has the following expression  (\cite{fazel2018global}, Lemma 1)
\begin{equation}
\label{eq:J_F_LQR}
  J(F) = \EE_{x_0 \sim \cD} \big[ x_0^\top P_F x_0 \big],   
\end{equation}
where $P_F$ is the solution to the following equation
\begin{equation}
\label{eq:P_F_equation}
P_F = Q_1 + F^\top R_1 F + (A-BF)^\top P_F (A-BF).  
\end{equation}
The gradient is given by
\#
\label{eq:gradient_J_F_LQR}
\nabla_F J(F) = 2 \Big( \big(R_1 + B^\top P_F B \big)F - B^\top P_F A\Big) \cdot \bigg[\EE_{x_0 \sim \cD} \sum_{t=0}^{\infty} x_t x_t^\top\bigg].
\#
Let $S_F = \sum_{t=0}^{\infty} x_t x_t^\top$ and observe that
\begin{equation}
\label{eq:S_F_equation}
S_F = x_0 x_0^\top + (A-BF)S_F(A-BF)^\top. 
\end{equation}
We start from some $F_0$ and apply our Algorithm~\ref{algo} to solve
the constrained LQR problem. In iteration $k$, with the current
estimator denoted by $F_k$, we first obtain an estimator of $P_{F_k}$
by starting from $Q_1$ and iteratively applying the recursion
$P_{F_k} \leftarrow Q_1 + F_k^\top R_1 F_k + (A-BF_k)^\top P_{F_k}
(A-BF_k)$ until convergence. Next, we sample an $x_0^*$ from the
distribution $\cD$ and follow a similar recursion given by
\eqref{eq:S_F_equation} to obtain an estimate of $S_{F_k}$.  Plugging
the sample $x_0^*$ and the estimates of $P_{F_k}$ and $S_{F_k}$ into
\eqref{eq:J_F_LQR} and \eqref{eq:gradient_J_F_LQR}, we obtain the
sample reward value $J^*(F_k)$ and $\nabla_F J^*(F_k)$,
respectively. With these two values, we follow
\eqref{eq:quadratic_appriximation_J} and \eqref{eq:J_bar} and obtain
$\overline J^{(k)}(F)$. We apply the same procedure to the cost
function $D(F)$ with $Q_1, R_1$ replaced by $Q_2, R_2$ to obtain
$\overline D^{(k)}(F)$. Finally we solve the optimization problem
\eqref{eq:QCQP} (or \eqref{eq:QCQP_alternative} if \eqref{eq:QCQP} is
infeasible) and obtain
$F_{k+1}$ by \eqref{eq:theta_update}.

\paragraph{Random state transition and control.} 
We then consider the setting where both $v_t$ and $w_t$ are
independent standard Gaussian white noise. In this case, the state
dynamic can be written as $x_{t+1} = (A-BF)x_t + \epsilon_t$ where
$\epsilon_t \sim \cN(0, I + BB^\top)$. Let $P_F$ be defined as
in \eqref{eq:P_F_equation} and $S_F$ be the solution to the following
Lyapunov equation
\begin{equation}
\label{eq:S_F_equation_2}
S_F = I + BB^\top + (A-BF)S_F(A-BF)^\top.   
\end{equation}
The objective function has the following expression (\cite{yang2019global}, Proposition 3.1) 
\#
\label{eq:J_F_LQR_2}
J(F) = \EE_{x \sim \cN(0, S_F)} \Big[ x^\top (Q_1+F^\top R_1 F) x  \Big] + \tr(R_1),
\#
and the gradient is given by
\begin{equation}
\label{eq:gradient_J_F_LQR_2}
\nabla_F J(F) = 2 \Big( \big(R_1 + B^\top P_F B \big)F - B^\top P_F A\Big) \cdot \EE_{x \sim \cN(0, S_F)}\Big[xx^\top \Big].
\end{equation}
Although in this setting it is straightforward to calculate the
expectation in a closed form, we keep the current expectation form to
be in line with our algorithm. Moreover, when the error distribution
is more complicated or unknown, we can no longer calculate the closed
form expression and have to sample in each iteration. With the
formulas given by \eqref{eq:J_F_LQR_2} and
\eqref{eq:gradient_J_F_LQR_2}, we again apply our
Algorithm~\ref{algo}. We sample $x \sim \cN(0, S_F)$ in each iteration
and solve  the optimization problem \eqref{eq:QCQP} or
\eqref{eq:QCQP_alternative}. The whole procedure is similar to the
random initial state case described above.

\paragraph{Other applications.} 
Our algorithm can also be applied to constrained parallel MDP and
constrained multi-agent MDP problem. Due to the space limit, we
relegate them to supplementary material.


\section{Experiment}
\label{sec:experiment}

We verify the effectiveness of the proposed algorithm through
experiments. We focus on the LQR setting with a random initial state as
discussed in Section \ref{sec:application_LQR}. In this experiment we
set $x \in \RR^{15}$ and $u \in \RR^{8}$. The initial state
distribution is uniform on the unit cube:
$x_0 \sim \cD = \text{Uniform} \big( [-1,1]^{15} \big)$. Each element
of $A$ and $B$ is sampled independently from the standard normal
distribution and scaled such that the eigenvalues of $A$ are within
the range $(-1,1)$. We initialize $F_0$ as an all-zero matrix, and the
choice of the constraint function and the value $D_0$ are such that
(1) the constrained problem is feasible; (2) the solution of the
unconstrained problem does not satisfy the constraint, i.e., the
problem is not trivial; (3) the initial value $F_0$ is not feasible.
The learning rates are set as $\eta_k = \frac{2}{3}k^{-3/4} $ and
$\rho_k = \frac{2}{3}k^{-2/3} $. The conservative choice of step size
is to avoid the situation where an eigenvalue of $A-BF$ runs out of
the range $(-1,1)$, and so the system is stable. 
\footnote{The code is available at \url{https://github.com/ming93/Safe_reinforcement_learning}}

\begin{figure}
    \centering
    \subfigure[Constraint value $D(\theta_k)$ in each iteration.]
    {
        \includegraphics[width=0.46 \textwidth]{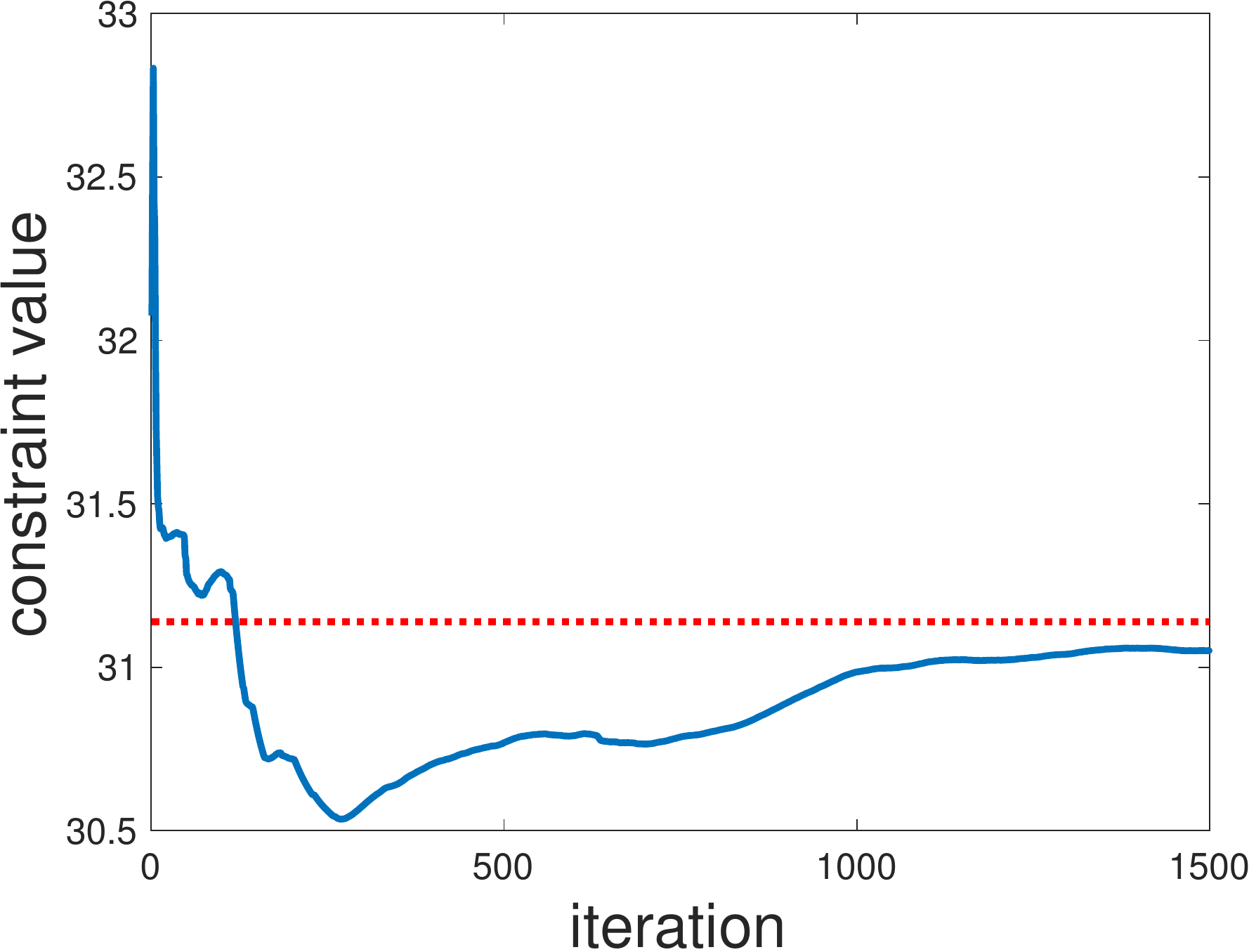}\hfill
        \label{fig:LQR_1_constraint}
    }
    \,
    \subfigure[Objective value $J(\theta_k)$ in each iteration.]
    {
        \includegraphics[width=0.46\textwidth]{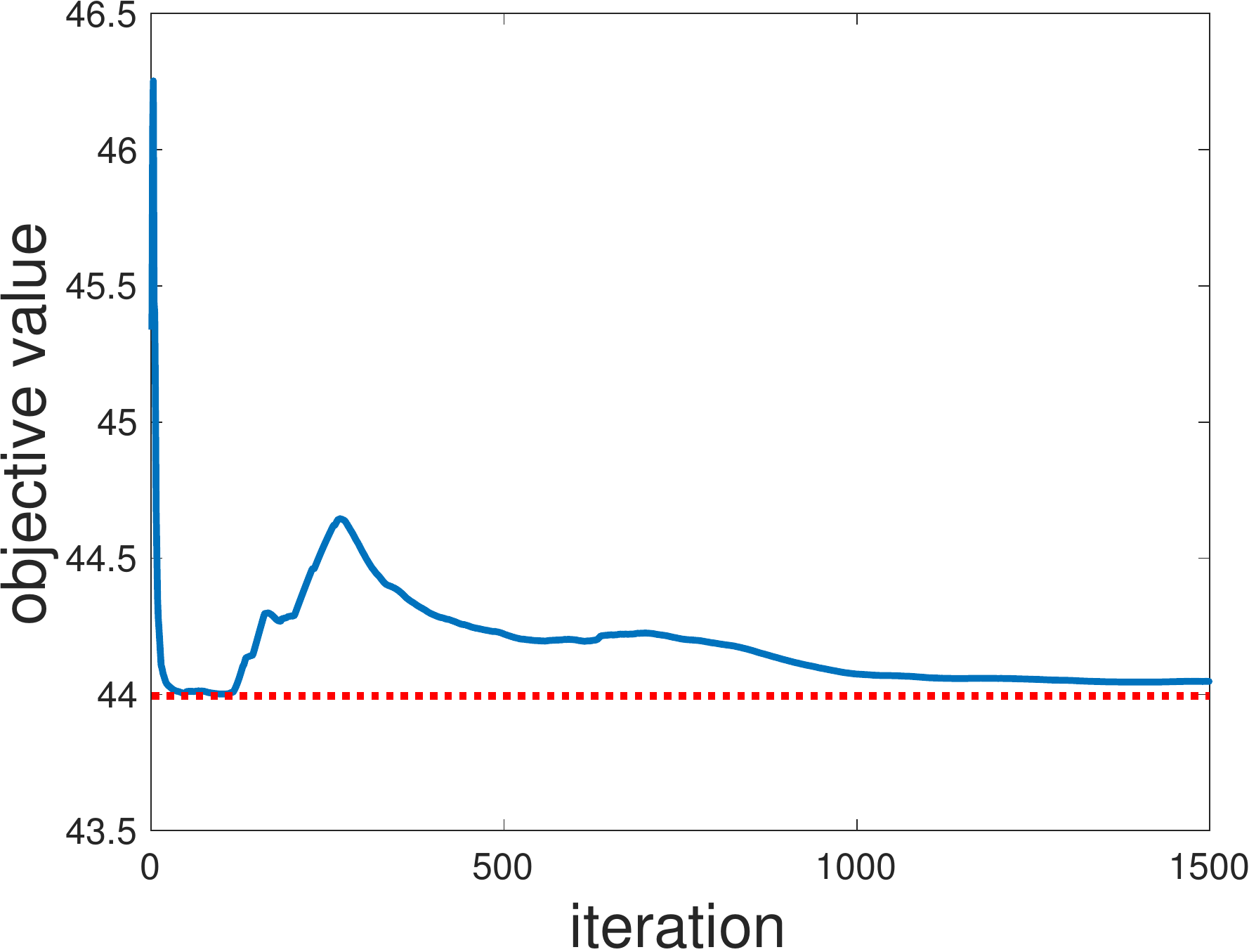}\hfill
        \label{fig:LQR_1_objective}
    }
    \caption{An experiment on constrained LQR problem. The iterate starts from an infeasible point and then becomes feasible and eventually converges.}
    \label{fig:LQR}
\end{figure}

Figure \ref{fig:LQR_1_constraint} and \ref{fig:LQR_1_objective} show
the constraint and objective value in each iteration, respectively.
The red horizontal line in Figure \ref{fig:LQR_1_constraint} is for
$D_0$, while the horizontal line in Figure \ref{fig:LQR_1_objective}
is for the unconstrained minimum objective value.  We can see from
Figure \ref{fig:LQR_1_constraint} that we start from an infeasible
point, and the problem becomes feasible after about 100
iterations. The objective value is in general decreasing after
becoming feasible, but never lower than the unconstrained minimum, as
shown in Figure \ref{fig:LQR_1_objective}.

\paragraph{Comparison with the Lagrangian method.} 
We  compare our proposed method with the usual Lagrangian method.
For the Lagrangian method, we follow the algorithm proposed in
\cite{chow2017risk} for safe reinforcement learning, which iteratively
applies gradient descent on the parameter $F$ and gradient ascent on
the Lagrangian multiplier $\lambda$ for the Lagrangian function until
convergence.  

Table \ref{table:experiment_compare_our_Lagrangian} reports the
comparison results with mean and standard deviation based on 50
replicates.  In the second and third columns, we compare the minimum
objective value and the number of iterations to achieve it.  We also
consider an approximate version, where we are satisfied with the
result if the objective value exceeds less than 0.02\% of the minimum
value. The fourth and fifth columns show the comparison results for
this approximate version.  We can see that both methods achieve
similar minimum objective values, but ours requires less number of
policy updates, for both minimum and approximate minimum version.


\begin{table}
\begin{center}
\begin{tabular}{c|cc|cc}
\hline
& min value & \verb|#| iterations & approx. min value & approx. \verb|#| iterations   \\\hline
Our method &   30.689 $\pm$ 0.114 & 2001 $\pm$ 1172 & 30.694 $\pm$ 0.114 & 604.3 $\pm$ 722.4\\ 
Lagrangian &    30.693 $\pm$ 0.113 & 7492 $\pm$ 1780 & 30.699 $\pm$ 0.113 & 5464 $\pm$ 2116  \\\hline
\end{tabular}
\end{center}
\caption{Comparison of our method with Lagrangian method}
\label{table:experiment_compare_our_Lagrangian}
\end{table}%



\clearpage
\bibliographystyle{plain}
\bibliography{paper}

\begin{thebibliography}{10}

\bibitem{achiam2017constrained}
Joshua Achiam, David Held, Aviv Tamar, and Pieter Abbeel.
\newblock Constrained policy optimization.
\newblock In {\em International Conference on Machine Learning}, pages 22--31,
  2017.

\bibitem{adolphs2018non}
Leonard Adolphs.
\newblock Non convex-concave saddle point optimization.
\newblock Master's thesis, ETH Zurich, 2018.

\bibitem{altman1999constrained}
Eitan Altman.
\newblock {\em Constrained Markov decision processes}, volume~7.
\newblock CRC Press, 1999.

\bibitem{ammar2015safe}
Haitham~Bou Ammar, Rasul Tutunov, and Eric Eaton.
\newblock Safe policy search for lifelong reinforcement learning with sublinear
  regret.
\newblock In {\em International Conference on Machine Learning}, pages
  2361--2369, 2015.

\bibitem{amodei2016concrete}
Dario Amodei, Chris Olah, Jacob Steinhardt, Paul Christiano, John Schulman, and
  Dan Man{\'e}.
\newblock Concrete problems in ai safety.
\newblock {\em arXiv preprint arXiv:1606.06565}, 2016.

\bibitem{anderson2007optimal}
Brian~DO Anderson and John~B Moore.
\newblock {\em Optimal control: linear quadratic methods}.
\newblock Courier Corporation, 2007.

\bibitem{bai2019provably}
Yu~Bai, Tengyang Xie, Nan Jiang, and Yu-Xiang Wang.
\newblock Provably efficient q-learning with low switching cost.
\newblock {\em arXiv preprint arXiv:1905.12849}, 2019.

\bibitem{bellemare2013arcade}
Marc~G Bellemare, Yavar Naddaf, Joel Veness, and Michael Bowling.
\newblock The arcade learning environment: An evaluation platform for general
  agents.
\newblock {\em Journal of Artificial Intelligence Research}, 47:253--279, 2013.

\bibitem{berkenkamp2017safe}
Felix Berkenkamp, Matteo Turchetta, Angela Schoellig, and Andreas Krause.
\newblock Safe model-based reinforcement learning with stability guarantees.
\newblock In {\em Advances in Neural Information Processing Systems}, pages
  908--918, 2017.

\bibitem{borkar1997stochastic}
Vivek~S Borkar.
\newblock Stochastic approximation with two time scales.
\newblock {\em Systems \& Control Letters}, 29(5):291--294, 1997.

\bibitem{boutilier1996planning}
Craig Boutilier.
\newblock Planning, learning and coordination in multiagent decision processes.
\newblock In {\em Proceedings of the 6th conference on Theoretical aspects of
  rationality and knowledge}, pages 195--210. Morgan Kaufmann Publishers Inc.,
  1996.

\bibitem{bradtke1993reinforcement}
Steven~J Bradtke.
\newblock Reinforcement learning applied to linear quadratic regulation.
\newblock In {\em Advances in neural information processing systems}, pages
  295--302, 1993.

\bibitem{bradtke1994adaptive}
Steven~J Bradtke, B~Erik Ydstie, and Andrew~G Barto.
\newblock Adaptive linear quadratic control using policy iteration.
\newblock In {\em Proceedings of the American control conference}, volume~3,
  pages 3475--3475. Citeseer, 1994.

\bibitem{breiman2001random}
Leo Breiman.
\newblock Random forests.
\newblock {\em Machine learning}, 45(1):5--32, 2001.

\bibitem{busoniu2008comprehensive}
Lucian Busoniu, Robert Babuska, and Bart De~Schutter.
\newblock A comprehensive survey of multiagent reinforcement learning.
\newblock {\em IEEE Transactions on Systems, Man, And Cybernetics-Part C:
  Applications and Reviews, 38 (2), 2008}, 2008.

\bibitem{cai2019neural}
Qi~Cai, Zhuoran Yang, Jason~D Lee, and Zhaoran Wang.
\newblock Neural temporal-difference learning converges to global optima.
\newblock {\em arXiv preprint arXiv:1905.10027}, 2019.

\bibitem{chen2018communication}
Tianyi Chen, Kaiqing Zhang, Georgios~B Giannakis, and Tamer Ba{\c{s}}ar.
\newblock Communication-efficient distributed reinforcement learning.
\newblock {\em arXiv preprint arXiv:1812.03239}, 2018.

\bibitem{chow2017risk}
Yinlam Chow, Mohammad Ghavamzadeh, Lucas Janson, and Marco Pavone.
\newblock Risk-constrained reinforcement learning with percentile risk
  criteria.
\newblock {\em Journal of Machine Learning Research}, 18(167):1--167, 2017.

\bibitem{chow2018lyapunov}
Yinlam Chow, Ofir Nachum, Edgar Duenez-Guzman, and Mohammad Ghavamzadeh.
\newblock A lyapunov-based approach to safe reinforcement learning.
\newblock {\em arXiv preprint arXiv:1805.07708}, 2018.

\bibitem{dann2014policy}
Christoph Dann, Gerhard Neumann, and Jan Peters.
\newblock Policy evaluation with temporal differences: A survey and comparison.
\newblock {\em The Journal of Machine Learning Research}, 15(1):809--883, 2014.

\bibitem{dean2017sample}
Sarah Dean, Horia Mania, Nikolai Matni, Benjamin Recht, and Stephen Tu.
\newblock On the sample complexity of the linear quadratic regulator.
\newblock {\em arXiv preprint arXiv:1710.01688}, 2017.

\bibitem{dunford1958linear}
Nelson Dunford and Jacob~T Schwartz.
\newblock {\em Linear operators part I: general theory}, volume~7.
\newblock Interscience publishers New York, 1958.

\bibitem{evans2005introduction}
Lawrence~C Evans.
\newblock An introduction to mathematical optimal control theory.
\newblock {\em Lecture Notes, University of California, Department of
  Mathematics, Berkeley}, 2005.

\bibitem{fazel2018global}
Maryam Fazel, Rong Ge, Sham Kakade, and Mehran Mesbahi.
\newblock Global convergence of policy gradient methods for the linear
  quadratic regulator.
\newblock In {\em International Conference on Machine Learning}, pages
  1466--1475, 2018.

\bibitem{fisac2018general}
Jaime~F Fisac, Anayo~K Akametalu, Melanie~N Zeilinger, Shahab Kaynama, Jeremy
  Gillula, and Claire~J Tomlin.
\newblock A general safety framework for learning-based control in uncertain
  robotic systems.
\newblock {\em IEEE Transactions on Automatic Control}, 2018.

\bibitem{garcia2015comprehensive}
Javier Garc{\i}a and Fernando Fern{\'a}ndez.
\newblock A comprehensive survey on safe reinforcement learning.
\newblock {\em Journal of Machine Learning Research}, 16(1):1437--1480, 2015.

\bibitem{goodfellow2014generative}
Ian Goodfellow, Jean Pouget-Abadie, Mehdi Mirza, Bing Xu, David Warde-Farley,
  Sherjil Ozair, Aaron Courville, and Yoshua Bengio.
\newblock Generative adversarial nets.
\newblock In {\em Advances in neural information processing systems}, pages
  2672--2680, 2014.

\bibitem{grondman2012survey}
Ivo Grondman, Lucian Busoniu, Gabriel~AD Lopes, and Robert Babuska.
\newblock A survey of actor-critic reinforcement learning: Standard and natural
  policy gradients.
\newblock {\em IEEE Transactions on Systems, Man, and Cybernetics, Part C
  (Applications and Reviews)}, 42(6):1291--1307, 2012.

\bibitem{he2019xbart}
Jingyu He, Saar Yalov, and P~Richard Hahn.
\newblock X{BART}: Accelerated {B}ayesian additive regression trees.
\newblock In {\em The 22nd International Conference on Artificial Intelligence
  and Statistics}, pages 1130--1138, 2019.

\bibitem{he2019scalable}
Jingyu He, Saar Yalov, Jared Murray, and P~Richard Hahn.
\newblock Stochastic tree ensembles for regularized supervised learning.
\newblock {\em Technical report}, 2019.

\bibitem{huang2018learning}
Jessie Huang, Fa~Wu, Doina Precup, and Yang Cai.
\newblock Learning safe policies with expert guidance.
\newblock {\em arXiv preprint arXiv:1805.08313}, 2018.

\bibitem{kelley2017general}
John~L Kelley.
\newblock {\em General topology}.
\newblock Courier Dover Publications, 2017.

\bibitem{konda2000actor}
Vijay~R Konda and John~N Tsitsiklis.
\newblock Actor-critic algorithms.
\newblock In {\em Advances in neural information processing systems}, pages
  1008--1014, 2000.

\bibitem{kretchmar2002parallel}
R~Matthew Kretchmar.
\newblock Parallel reinforcement learning.
\newblock In {\em The 6th World Conference on Systemics, Cybernetics, and
  Informatics}. Citeseer, 2002.

\bibitem{lacotte2018risk}
Jonathan Lacotte, Yinlam Chow, Mohammad Ghavamzadeh, and Marco Pavone.
\newblock Risk-sensitive generative adversarial imitation learning.
\newblock {\em arXiv preprint arXiv:1808.04468}, 2018.

\bibitem{lee2018modular}
Dennis Lee, Haoran Tang, Jeffrey~O Zhang, Huazhe Xu, Trevor Darrell, and Pieter
  Abbeel.
\newblock Modular architecture for starcraft ii with deep reinforcement
  learning.
\newblock In {\em Fourteenth Artificial Intelligence and Interactive Digital
  Entertainment Conference}, 2018.

\bibitem{li2017deep}
Yuxi Li.
\newblock Deep reinforcement learning: An overview.
\newblock {\em arXiv preprint arXiv:1701.07274}, 2017.

\bibitem{liu2018stochastic}
An~Liu, Vincent Lau, and Borna Kananian.
\newblock Stochastic successive convex approximation for non-convex constrained
  stochastic optimization.
\newblock {\em arXiv preprint arXiv:1801.08266}, 2018.

\bibitem{liu2019neural}
Boyi Liu, Qi~Cai, Zhuoran Yang, and Zhaoran Wang.
\newblock Neural proximal/trust region policy optimization attains globally
  optimal policy.
\newblock {\em arXiv preprint arXiv:1906.10306}, 2019.

\bibitem{mnih2016asynchronous}
Volodymyr Mnih, Adria~Puigdomenech Badia, Mehdi Mirza, Alex Graves, Timothy
  Lillicrap, Tim Harley, David Silver, and Koray Kavukcuoglu.
\newblock Asynchronous methods for deep reinforcement learning.
\newblock In {\em International conference on machine learning}, pages
  1928--1937, 2016.

\bibitem{mnih2015human}
Volodymyr Mnih, Koray Kavukcuoglu, David Silver, Andrei~A Rusu, Joel Veness,
  Marc~G Bellemare, Alex Graves, Martin Riedmiller, Andreas~K Fidjeland, Georg
  Ostrovski, et~al.
\newblock Human-level control through deep reinforcement learning.
\newblock {\em Nature}, 518(7540):529, 2015.

\bibitem{nair2015massively}
Arun Nair, Praveen Srinivasan, Sam Blackwell, Cagdas Alcicek, Rory Fearon,
  Alessandro De~Maria, Vedavyas Panneershelvam, Mustafa Suleyman, Charles
  Beattie, Stig Petersen, et~al.
\newblock Massively parallel methods for deep reinforcement learning.
\newblock {\em arXiv preprint arXiv:1507.04296}, 2015.

\bibitem{paternain2018stochastic}
Santiago Paternain.
\newblock {\em Stochastic Control Foundations of Autonomous Behavior}.
\newblock PhD thesis, University of Pennsylvania, 2018.

\bibitem{peng2017multiagent}
Peng Peng, Ying Wen, Yaodong Yang, Quan Yuan, Zhenkun Tang, Haitao Long, and
  Jun Wang.
\newblock Multiagent bidirectionally-coordinated nets: Emergence of human-level
  coordination in learning to play starcraft combat games.
\newblock {\em arXiv preprint arXiv:1703.10069}, 2017.

\bibitem{pirotta2015policy}
Matteo Pirotta, Marcello Restelli, and Luca Bascetta.
\newblock Policy gradient in lipschitz markov decision processes.
\newblock {\em Machine Learning}, 100(2-3):255--283, 2015.

\bibitem{prashanth2016variance}
LA~Prashanth and Mohammad Ghavamzadeh.
\newblock Variance-constrained actor-critic algorithms for discounted and
  average reward mdps.
\newblock {\em Machine Learning}, 105(3):367--417, 2016.

\bibitem{recht2018tour}
Benjamin Recht.
\newblock A tour of reinforcement learning: The view from continuous control.
\newblock {\em Annual Review of Control, Robotics, and Autonomous Systems},
  2018.

\bibitem{rhee2015unbiased}
Chang-han Rhee and Peter~W Glynn.
\newblock Unbiased estimation with square root convergence for sde models.
\newblock {\em Operations Research}, 63(5):1026--1043, 2015.

\bibitem{ruszczynski1980feasible}
Andrzej Ruszczy{\'n}ski.
\newblock Feasible direction methods for stochastic programming problems.
\newblock {\em Mathematical Programming}, 19(1):220--229, 1980.

\bibitem{scharpff2016solving}
Joris Scharpff, Diederik~M Roijers, Frans~A Oliehoek, Matthijs~TJ Spaan, and
  Mathijs~Michiel de~Weerdt.
\newblock Solving transition-independent multi-agent mdps with sparse
  interactions.
\newblock In {\em AAAI}, pages 3174--3180, 2016.

\bibitem{schulman2015trust}
John Schulman, Sergey Levine, Pieter Abbeel, Michael Jordan, and Philipp
  Moritz.
\newblock Trust region policy optimization.
\newblock In {\em International Conference on Machine Learning}, pages
  1889--1897, 2015.

\bibitem{scutari2013decomposition}
Gesualdo Scutari, Francisco Facchinei, Peiran Song, Daniel~P Palomar, and
  Jong-Shi Pang.
\newblock Decomposition by partial linearization: Parallel optimization of
  multi-agent systems.
\newblock {\em IEEE Transactions on Signal Processing}, 62(3):641--656, 2013.

\bibitem{silver2016mastering}
David Silver, Aja Huang, Chris~J Maddison, Arthur Guez, Laurent Sifre, George
  Van Den~Driessche, Julian Schrittwieser, Ioannis Antonoglou, Veda
  Panneershelvam, Marc Lanctot, et~al.
\newblock Mastering the game of go with deep neural networks and tree search.
\newblock {\em nature}, 529(7587):484, 2016.

\bibitem{silver2018general}
David Silver, Thomas Hubert, Julian Schrittwieser, Ioannis Antonoglou, Matthew
  Lai, Arthur Guez, Marc Lanctot, Laurent Sifre, Dharshan Kumaran, Thore
  Graepel, et~al.
\newblock A general reinforcement learning algorithm that masters chess, shogi,
  and go through self-play.
\newblock {\em Science}, 362(6419):1140--1144, 2018.

\bibitem{silver2017mastering}
David Silver, Julian Schrittwieser, Karen Simonyan, Ioannis Antonoglou, Aja
  Huang, Arthur Guez, Thomas Hubert, Lucas Baker, Matthew Lai, Adrian Bolton,
  et~al.
\newblock Mastering the game of go without human knowledge.
\newblock {\em Nature}, 550(7676):354, 2017.

\bibitem{sun2018tstarbots}
Peng Sun, Xinghai Sun, Lei Han, Jiechao Xiong, Qing Wang, Bo~Li, Yang Zheng,
  Ji~Liu, Yongsheng Liu, Han Liu, et~al.
\newblock Tstarbots: Defeating the cheating level builtin ai in starcraft ii in
  the full game.
\newblock {\em arXiv preprint arXiv:1809.07193}, 2018.

\bibitem{sun2016majorization}
Ying Sun, Prabhu Babu, and Daniel~P Palomar.
\newblock Majorization-minimization algorithms in signal processing,
  communications, and machine learning.
\newblock {\em IEEE Transactions on Signal Processing}, 65(3):794--816, 2016.

\bibitem{sutton2018reinforcement}
Richard~S Sutton and Andrew~G Barto.
\newblock {\em Reinforcement learning: An introduction}.
\newblock MIT press, 2018.

\bibitem{sutton2000policy}
Richard~S Sutton, David~A McAllester, Satinder~P Singh, and Yishay Mansour.
\newblock Policy gradient methods for reinforcement learning with function
  approximation.
\newblock In {\em Advances in neural information processing systems}, pages
  1057--1063, 2000.

\bibitem{tessler2018reward}
Chen Tessler, Daniel~J Mankowitz, and Shie Mannor.
\newblock Reward constrained policy optimization.
\newblock {\em arXiv preprint arXiv:1805.11074}, 2018.

\bibitem{turchetta2016safe}
Matteo Turchetta, Felix Berkenkamp, and Andreas Krause.
\newblock Safe exploration in finite markov decision processes with gaussian
  processes.
\newblock In {\em Advances in Neural Information Processing Systems}, pages
  4312--4320, 2016.

\bibitem{vinyals2017starcraft}
Oriol Vinyals, Timo Ewalds, Sergey Bartunov, Petko Georgiev, Alexander~Sasha
  Vezhnevets, Michelle Yeo, Alireza Makhzani, Heinrich K{\"u}ttler, John
  Agapiou, Julian Schrittwieser, et~al.
\newblock Starcraft ii: A new challenge for reinforcement learning.
\newblock {\em arXiv preprint arXiv:1708.04782}, 2017.

\bibitem{wai2018multi}
Hoi-To Wai, Zhuoran Yang, Zhaoran Wang, and Mingyi Hong.
\newblock Multi-agent reinforcement learning via double averaging primal-dual
  optimization.
\newblock {\em arXiv preprint arXiv:1806.00877}, 2018.

\bibitem{wang2019neural}
Lingxiao Wang, Qi~Cai, Zhuoran Yang, and Zhaoran Wang.
\newblock Neural policy gradient methods: Global optimality and rates of
  convergence.
\newblock {\em arXiv preprint arXiv:1909.01150}, 2019.

\bibitem{wen2018constrained}
Min Wen and Ufuk Topcu.
\newblock Constrained cross-entropy method for safe reinforcement learning.
\newblock In {\em Advances in Neural Information Processing Systems}, pages
  7461--7471, 2018.

\bibitem{xu2018macro}
Sijia Xu, Hongyu Kuang, Zhi Zhuang, Renjie Hu, Yang Liu, and Huyang Sun.
\newblock Macro action selection with deep reinforcement learning in starcraft.
\newblock {\em arXiv preprint arXiv:1812.00336}, 2018.

\bibitem{yang2016parallel}
Yang Yang, Gesualdo Scutari, Daniel~P Palomar, and Marius Pesavento.
\newblock A parallel decomposition method for nonconvex stochastic multi-agent
  optimization problems.
\newblock {\em IEEE Transactions on Signal Processing}, 64(11):2949--2964,
  2016.

\bibitem{yang2019global}
Zhuoran Yang, Yongxin Chen, Mingyi Hong, and Zhaoran Wang.
\newblock On the global convergence of actor-critic: A case for linear
  quadratic regulator with ergodic cost.
\newblock {\em arXiv preprint arXiv:1907.06246}, 2019.

\bibitem{zhang2018finite}
Kaiqing Zhang, Zhuoran Yang, Han Liu, Tong Zhang, and Tamer Ba{\c{s}}ar.
\newblock Finite-sample analyses for fully decentralized multi-agent
  reinforcement learning.
\newblock {\em arXiv preprint arXiv:1812.02783}, 2018.

\end{thebibliography}

\clearpage
\appendix

\section{Other applications}
\label{sec:other_applications}

\subsection{Constrained Parallel Markov Decision Process}

We consider the parallel MDP problem \cite{kretchmar2002parallel,
  nair2015massively, chen2018communication} where we have a
single-agent MDP task and $N$ workers, where each worker acts as an
individual agent and aims to solve the \emph{same} MDP problem.  In
the parallel MDP setting, each agent is characterized by a tuple
$(\cS, \cA, P, \gamma, r^i, d^i, \mu^i)$, where each agent has the
same but individual state space, action space, transition probability
distribution, and the discount factor.  However, the reward
function, cost function, and the distribution of the initial state
$s_0\in \cS$ could be different for each agent, but satisfy
$\EE[r^i(s,a)] = r(s,a)$, $\EE[d^i(s,a)] = d(s,a)$,
and $\EE[\mu^i(s,a)] = \mu(s,a)$. Each agent $i$ generates its own
trajectory $\{s^i_0, a^i_0, s^i_1, a^i_1, \dots\}$ and collects its own
reward/cost value $\{r^i_0, d^i_0, r^i_1, d^i_1, \dots \}$.

The hope is that by solving the single-agent problem using $N$ agents
in parallel, the algorithm could be more stable and converge much
faster \cite{mnih2016asynchronous}. Intuitively, each agent $i$ may
have a different initial state and will explore different parts of the
state space due to the randomness in the state transition distribution
and the policy.  It also helps to reduce the correlation between
agents' behaviors.  As a result, by running multiple agents in
parallel, we are more likely to visit different parts of the
environment and get the experience of the reward/cost function values
more efficiently. This mimics the strategy used in tree-based supervised learning algorithms \cite{breiman2001random, he2019xbart, he2019scalable}. 

Following the settings in \cite{chen2018communication}, we have $N$
agents (i.e., $N$ workers) and one central controller in the
system. The global parameter is denoted by $\theta$, and we consider
the constrained parallel MDP problem where the goal is to solve the
following optimization problem:
\begin{equation}
\begin{aligned}
\label{eq:constrained_parallel_MDP}
&\mathop{\textrm{minimize}}_{\theta} 
~~J(\theta) =  \sum_{i=1}^N  \EE_{\pi_{\theta} } \biggl [ - \sum_{t\geq 0} \gamma ^t \cdot r^i(s^i_t, a^i_t) \biggr ], \\
&\text{subject to} 
~~D (\theta) =   \EE_{\pi_{\theta} } \biggl [  \sum_{t\geq 0} \gamma ^t \cdot  d^i(s^i_t, a^i_t) \biggr ]  \leq  D_0, 
~~i \in \cN.
\end{aligned}
\end{equation}
During the estimation step, the controller broadcasts the current
parameter $\theta_k$ to each agent and each agent samples its own
trajectory and obtains estimators for function value/gradient of the
reward/cost function. Next, each agent uploads its estimators to the
central controller and the central controller takes the average over
these estimators, and then follow our proposed algorithm to solve for
the QCQP problem and update the parameter to $\theta_{k+1}$. This
process continues until convergence.

\subsection{Constrained Multi-agent Markov Decision Process}

A natural extension of the (single-agent) MDP is to consider a model
with $N$ agents termed multi-agent Markov decision process (MMDP).
Recently this kind of problem has been attracting more and more
attention.  See \cite{busoniu2008comprehensive} for a comprehensive
survey.  Most of the work on multi-agent MDP problems consider the
setting where the agents share the same global state space, but each
with their own collection of actions and rewards
\cite{boutilier1996planning, wai2018multi, zhang2018finite}.  In each
stage of the system, each agent observes the global state and chooses
its own action individually. As a result, each agent receives its
reward and the state evolves according to the joint transition
distribution.  An MMDP problem can be fully collaborative where all
the agents have the same goal, or fully competitive where the problem
consists of two agents with an opposite goal, or the mix of the two.

Here we consider a slightly different setting where each agent has its
own state space.  The only connection between the agents is that the
global reward is a function of the overall states and actions.
Furthermore, each agent has its own constraint which depends on its
own state and action only.  This problem is known as
Transition-Independent Multi-agent MDP and is considered in
\cite{scharpff2016solving}.  Specifically, each agent's task is
characterized by a tuple $(\cS^i, \cA^i, P^i, \gamma, d^i, \mu^i)$
with each component defined as usual.  Note that
$P^i \colon \cS^i \times \cA^i \rightarrow \cP(\cS^i)$ and
$d^i \colon \cS^i \times \cA^i \rightarrow \RR$ are functions of each
agent's state and action only and do not depend on other agents.
Denote $\cS = \Pi_{i \in \cN} \cS^i$ and $\cA = \Pi_{i \in \cN} \cA^i$
as the joint state space and action space. The global reward function
is given by $r \colon \cS \times \cA \rightarrow \RR$ that depends on
the joint state and action.  Here we consider the fully collaborative
setting where all the agents have the same goal.
Under this setting, the policy set of each agent is parameterized as
$\{ \pi^i_{\theta^i} \colon \cS^i \rightarrow \cP(\cA^i)\}$ and we
denote $\theta = [\theta^1, \ldots, \theta^N]$ as the overall
parameters and $\pi_{\theta}$ as the overall policy.
In the following, we use $\cN = \{1, 2, \ldots, N\}$ to denote the $N$
agents.
Denote $a^i_t$ as the action chosen by agent $i$ at stage $t$ and
$a_t = \Pi_{i \in \cN} ~ a^i_t$ as the joint action chosen by all the
agents.  The goal of this constrained MMDP is to solve the following
problem
\begin{equation}
\begin{aligned}
\label{eq:constrained_MMDP}
&\mathop{\textrm{minimize}}_{\theta} 
~~J(\theta) =  \EE_{\pi_{\theta} } \biggl [ - \sum_{t\geq 0} \gamma ^t \cdot r(s_t, a_t) \biggr ], \\
&\text{subject to} 
~~D^i (\theta^i) =   \EE_{\pi_{\theta^i} } \biggl [  \sum_{t\geq 0} \gamma ^t \cdot  d^i(s^i_t, a^i_t) \biggr ]  \leq  D^i_0, 
~~i \in \cN.
\end{aligned}
\end{equation}
Inspired by the parallel implementation (\cite{liu2018stochastic},
Section V), our algorithm applies naturally to constrained MMDP
problem with some modifications.  This modified procedure can also be
viewed as a distributed version of the original algorithm.  The
overall problem \eqref{eq:constrained_MMDP} can be viewed as a large
``single-agent" problem where the constraints are decomposable into
$N$ parts. In this case, instead of solving a large QCQP problem in
each iteration, each agent could solve its own QCQP problem in a
distributed manner which is much more efficient.  As before, we denote
the sample negative reward and cost function as
\begin{equation*}
   J^*(\theta) = -
\sum_{t\geq 0} \gamma ^t \cdot r(s_t, a_t) \qquad \text{and}\qquad
D^{i,*}(\theta^i) = \sum_{t\geq 0} \gamma ^t \cdot d^i(s^i_t, a^i_t).
\end{equation*}
In each iteration with $\theta_k = [\theta^1_k, ..., \theta^N_k]$, we approximate $J(\theta)$ and $D(\theta)$ as 
\begin{align}
\label{eq:quadratic_appriximation_J_MMDP}
\tilde J^i(\theta^i, \theta_k, \tau ) & = \frac{1}{N} J^{*}(\theta_k) + \la \nabla_{\theta^i} J^{*}(\theta_k), \theta^i - \theta^i_k \ra + \tau \| \theta^i - \theta^i_k \|_2^2,  \\
\label{eq:quadratic_appriximation_MMDP}
\tilde D^{i}(\theta^i, \theta_k, \tau ) &= D^{i,*}(\theta^i_k)  +  \la \nabla _{\theta^i} D^{i,*}(\theta^i_k), \theta^i - \theta^i_k \ra + \tau  \| \theta^i - \theta^i_k \|_2^2.  
\end{align}
Note that the constraint function is naturally decomposable into $N$
parts. We also ``manually" split the objective function into $N$
parts, so that each agent could solve its own QCQP problem in a
distributed manner.  As before, we define
\begin{align*}
\overline J^{i, (k)} (\theta^i) = (1 - \rho_k) \cdot \overline J^{i,(k-1)} (\theta^i) + \rho_k \cdot \tilde J^i(\theta^i, \theta_k, \tau), \\
\overline D^{i, (k)} (\theta^i) = (1 - \rho_k) \cdot \overline
D^{i,(k-1)} (\theta^i) + \rho_k \cdot \tilde D^i(\theta^i, \theta_k,
\tau).   
\end{align*}
With this surrogate functions, each agent then solves its own convex
relaxation problem
\begin{equation}
\label{eq:QCQP_MMDP}
\overline \theta^i_k = \argmin _{\theta^i }~ \overline J^{i,(k)} (\theta^i) \qquad \text{subject to}\qquad \overline  D^{i,(k)} (\theta^i) \leq D^i_0,  
\end{equation}
or, alternatively, solves for the feasibility problem if
\eqref{eq:QCQP_MMDP} is infeasible
\begin{equation}
\label{eq:QCQP_alternative_MMDP}
\overline \theta^i_k = \argmin _{\theta^i, \alpha^i }~ \alpha^i \qquad \text{subject to}\qquad \overline  D^{i,(k)} (\theta^i) \leq D^i_0 + \alpha^i.
\end{equation}
This step can be implemented in a distributed manner for each agent and
is more efficient than solving the overall problem with the overall
parameter $\theta$. Finally, the update rule for each agent $i$ is as
usual
\[
\theta^i_{t+1} = ( 1- \eta_k) \cdot \theta^i_k + \eta_k \cdot \overline \theta^i_k.
\]
This process continues until convergence.


\section{Proof of Theorem \ref{thm:main}}
\label{sec:proof}

According to the choice of the surrogate function
\eqref{eq:quadratic_appriximation_J} and Assumption
\ref{assumption:bounded}, it is straightforward to verify that the
function $\overline J^{(k)} (\theta)$ defined in \eqref{eq:J_bar} is
uniformly strongly convex in $\theta$ for each iteration
$t$. Moreover, both $\overline J^{(k)} (\theta)$ and
$\nabla_{\theta} \overline J^{(k)} (\theta)$ are Lipschitz continuous
functions.

From Lemma 1 in \cite{ruszczynski1980feasible} we have
\begin{equation*}
\lim_{t \to \infty} \Big| \overline J^{(k)} (\theta) - \EE \big[ \tilde J(\theta, \theta_k, \tau ) \big] \Big| = 0.  
\end{equation*}
Since the function $\EE \big[ \tilde J(\theta, \theta_k, \tau ) \big]$
is Lipschitz continuous in $\theta_k$, we obtain that
\[
\Big| \overline J^{(k_1)} (\theta) - \overline J^{(k_2)} (\theta) \Big| \leq L_0 \cdot \| \theta_{k_1} - \theta_{k_2} \| + \epsilon,
\]
for some constant $L_0$ and the error term $\epsilon$ that goes to
$0$ as $k_1,k_2$ go to infinity. This shows that the function
sequence $\overline J^{(k_j)} (\theta)$ is equicontinuous.  Since
$\Theta$ is compact and the discounted cumulative reward function is
bounded by $r_{\max}/(1-\gamma)$, we can apply Arzela-Ascoli theorem
\cite{dunford1958linear, kelley2017general} to prove existence of
$\hat J(\theta)$ that converges uniformly. Moreover, since we apply
the same operations on the constraint function $D(\theta)$ as to the
reward function $J(\theta)$ in Algorithm~\ref{algo}, the above
properties also hold for $D(\theta)$.

The rest of the proof follows in a similar way as the proof of Theorem
1 in \cite{liu2018stochastic}.  Under
Assumptions~\ref{assumption:step_size} - \ref{assumption:feasible},
the technical conditions in \cite{liu2018stochastic} are satisfied by
the choice of the surrogate functions
\eqref{eq:quadratic_appriximation_J} and
\eqref{eq:quadratic_appriximation_D}. According to Lemma 2 in
\cite{liu2018stochastic}, with probability one we have
\[
\mathop{\lim\sup}_{k \to \infty} D(\theta_k) \leq D_0.
\]
This shows that, although in some of the iterations the convex
relaxation problem \eqref{eq:QCQP} is infeasible, and we have to solve
the alternative problem \eqref{eq:QCQP_alternative}, the iterates
$\{\theta_k\}$ converge to the feasible region of the original
problem \eqref{eq:constrained} with probability one. Furthermore, with
probability one, the convergent point $\tilde \theta$ is the optimal
solution to the following problem
\begin{equation}
\begin{aligned}
\label{eq:problem_converged}
\mathop{\textrm{minimize}}_{\theta \in \Theta} 
~~ \hat J(\theta)  \qquad
\text{subject to} \qquad
 \hat D(\theta)   \leq  D_0.
\end{aligned}
\end{equation}
The KKT conditions for \eqref{eq:problem_converged} together with the
Slater condition show that the KKT conditions of the original problem
\eqref{eq:constrained} are also satisfied at $\tilde \theta$. This
shows that $\tilde \theta$ is a stationary point of the original
problem almost surely.


\end{document}